\documentclass[letterpaper, 10 pt, journal, twoside]{IEEEtran}

\usepackage{amsmath,amsfonts}
\usepackage{algorithmic}
\usepackage{algorithm}
\usepackage{array}
\usepackage[caption=false,font=footnotesize,labelfont=rm,textfont=rm]{subfig}  
\usepackage{textcomp}
\usepackage{stfloats}
\usepackage{url}
\usepackage{verbatim}
\usepackage{graphicx}
\usepackage{cite}
\usepackage{amssymb}
\usepackage{diagbox}
\usepackage{float}
\usepackage{color}
\usepackage{xspace}
\usepackage{booktabs}
\usepackage{bbding,tikz}
\usepackage{multirow,multicol}
\usepackage{textcase}
\usepackage{threeparttable}
\usepackage{tabu}
\usepackage{flushend}
\usepackage{makecell}

\usepackage{changes}

\DeclareSubrefFormat{subsimple}{#1(#2)}

\makeatletter
\let\NAT@parse\undefined
\makeatother
\usepackage[colorlinks,
            urlcolor=blue,
            linkcolor=blue,   
            anchorcolor=blue, 
            citecolor=green   
            ]{hyperref}

\usepackage{tikz,xcolor,hyperref}
\definecolor{lime}{HTML}{A6CE39}
\DeclareRobustCommand{\orcidicon}{%
    \begin{tikzpicture}
    \draw[lime, fill=lime] (0,0)
    circle [radius=0.16]
    node[white] {{\fontfamily{qag}\selectfont \tiny ID}}; \draw[white, fill=white] (-0.0625,0.095) 
    circle [radius=0.007]; \end{tikzpicture}
    \hspace{-2mm}}
\foreach \x in {A, ..., Z}{%
    \expandafter\xdef\csname orcid\x\endcsname{\noexpand\href{https://orcid.org/\csname orcidauthor\x\endcsname}{\noexpand\orcidicon}}
    }

\begin{document}

\newcommand{\core}{mmPlace\xspace}
\newcommand{\orcidauthorA}{0009-0006-4285-1905}
\newcommand{\orcidauthorB}{0000-0001-9046-798X}

\newcommand{\ynote}[1]{{ \color{red}    [Yanyong:   #1] \color{black}}}
\newcommand{\dnote}[1]{{ \color{purple} [Yifan:     #1] \color{black}}}
\newcommand{\mnote}[1]{{ \color{blue}   [Chengzhen: #1] \color{black}}}

\title{\core: Robust Place Recognition with Intermediate Frequency Signal of Low-cost Single-chip Millimeter Wave Radar
}

\author{Chengzhen Meng\orcidA{}, Yifan Duan, Chenming He, Dequan Wang, \\ Xiaoran Fan, and Yanyong Zhang\orcidB{}, \emph{Fellow, IEEE}
\thanks{Manuscript received: November, 4, 2023; Revised February, 3, 2024; Accepted March, 5, 2024. This paper was recommended for publication by Editor S. Behnke upon evaluation of the reviewers' comments. This work was supported by the National Natural Science Foundation of China (No.62332016). \emph{(Corresponding author: Yanyong Zhang.)}}
\thanks{Chengzhen Meng, Yifan Duan, Chenming He, Dequan Wang, and Yanyong Zhang are with the School of Computer Science and Technology, University of Science and Technology of China, Hefei, China
       {\tt\footnotesize (e-mail: \{czmeng, dyf0202, hechenming, wdq15588\}@mail.ustc.edu.cn, \{yanyongz\}@ustc.edu.cn)}.}
\thanks{Xiaoran Fan is with the Technology Directions Office (TDO) of Google
       {\tt\footnotesize (e-mail: ox5bc@winlab.rutgers.edu)}. }
\thanks{Digital Object Identifier (DOI): see top of this page.}
}

\markboth{IEEE Robotics and Automation Letters. Preprint Version. Accepted March, 2024}
{Meng \MakeLowercase{\textit{et al.}}: \core: Robust Place Recognition with Low-cost Single-chip Millimeter Wave Radar} 

\maketitle

\begin{abstract}
Place recognition is crucial for tasks like loop-closure detection and re-localization. Single-chip millimeter wave radar (single-chip radar in short) emerges as a low-cost sensor option for place recognition, with the advantage of insensitivity to degraded visual environments. However, it encounters two challenges. Firstly, sparse point cloud from single-chip radar leads to poor performance when using current place recognition methods, which assume much denser data. Secondly, its performance significantly declines in scenarios involving rotational and lateral variations, due to limited overlap in its field of view (FOV). We propose \core, a robust place recognition system to address these challenges. Specifically, \core transforms intermediate frequency (IF) signal into range azimuth heatmap and employs a spatial encoder to extract features. Additionally, to improve the performance in scenarios involving rotational and lateral variations, \core employs a rotating platform and concatenates heatmaps in a rotation cycle, effectively expanding the system's FOV. We evaluate \core's performance on the milliSonic dataset, which is collected on the University of Science and Technology of China (USTC) campus, the city roads surrounding the campus, and an underground parking garage. The results demonstrate that \core outperforms point cloud-based methods and achieves 87.37\% recall@1 in scenarios involving rotational and lateral variations.
\end{abstract}

\section{INTRODUCTION}

\IEEEPARstart{P}{lace} recognition plays a vital role in various fields~\cite{yin2022general}, such as robotics, autonomous vehicles, augmented reality, and more. The main goal of place recognition is to identify previously visited locations based on sensor data and match them with a pre-built map database. For example, in Simultaneous Localization and Mapping (SLAM), place recognition plays an essential role in loop-closure detection, which helps correct the accumulated error in the robot's estimated trajectory. Moreover, in long-term navigation, place recognition assists with re-localization, enabling the robot to determine its position within the map after an extended period or after being temporarily lost.

\begin{figure}[t]
    \centering
    \setlength{\abovecaptionskip}{0.cm}
    \includegraphics[width=1\linewidth]{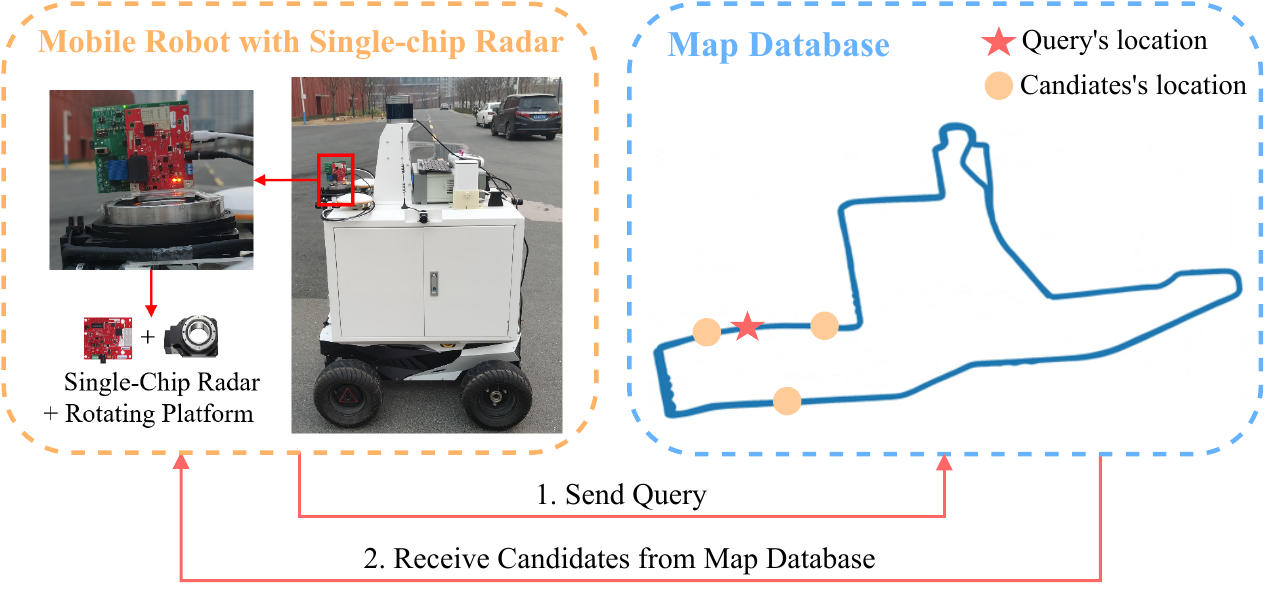}
    \caption{Our single-chip radar place recognition system. It finds the same location in a pre-built map database based on the given query data.}
    \label{fig:place_regcognition}
    \vspace{-10pt}
\end{figure}

\begin{table}[t]

\renewcommand\arraystretch{1.2} 
\setlength{\tabcolsep}{1.5mm}   

\centering

\setlength{\abovecaptionskip}{0cm}
\caption{Comparison of Different Radar Place Recognition Systems}
\label{tab:table_radar_type}

\begin{threeparttable}
\begin{tabular}{c|c|c|c|c}
\Xhline{1.2pt}
System & Oxford~\cite{barnes2020oxford} & nuScenes~\cite{caesar2020nuscenes} & Ours w/o RP\tnote{1} & \textbf{Ours w/ RP\tnote{2}} \\ \hline
Type            & mechanical & single-chip     & single-chip & \textbf{single-chip}  \\
Product         & CTS350-X   & ARS408$\times$5 & AWR1642     & \textbf{AWR1642$+$RP} \\
Cost(\$)        & 40000      & 520$\times$5    & 299         & \textbf{299$+$90}     \\
Point           & 1000       & 250$\times$5    & 200         & \textbf{200}          \\
FOV($^{\circ}$) & 360        & 360             & 120         & \textbf{300}          \\
\Xhline{1.2pt}
\end{tabular}

\begin{tablenotes}
    \footnotesize
    \item[1] w/o RP: without the rotating platform.
    \item[2] w/ RP: with the rotating platform.
\end{tablenotes}
\end{threeparttable}
    
\vspace{-15pt}

\end{table}

Camera and LiDAR are currently the dominant sensors for place recognition~\cite{lowry2015visual}. However, due to they are both optical sensors, their performance degrades severely in degraded visual environments, such as fog, rain, and snow~\cite{paek2022k, cai2021autoplace, suaftescu2020kidnapped}. On the other hand, the radar exhibits insensitivity to degraded visual environments due to its longer (than vision) wavelength ($\lambda \approx 4mm$). Specifically, the radar utilizes the millimeter-wave signal with a wavelength larger than the small particles in fog, rain, and snow, enabling easy pass through raindrops and snowflakes~\cite{yin2022general, paek2022k, qian2021robust}.

Currently, radars used in place recognition systems~\cite{harlow2023new} can be categorized into two types: mechanical and single-chip. As shown in Tab.~\ref{tab:table_radar_type}, although single-chip radar has fewer point clouds and a smaller FOV than mechanical radar, it offers the advantage of being much more affordable. However, leveraging a single-chip radar for place recognition presents two following challenges. 
Firstly, the point cloud data of a single-chip radar is sparse. Consequently, it encounters performance degradation when incorporated into current place recognition methods~\cite{kim2018scan, kim2021scan, vidanapathirana2022logg3d, hong2020radarslam, barnes2020under} that rely on the dense point cloud.
Secondly, the restricted FOV of the single-chip radar exhibits limited overlap between the current query data and the candidate data stored in the pre-built map database, particularly in scenarios with rotational and lateral variations. Consequently, the performance of the single-chip radar place recognition noticeably degrades in scenarios involving rotational and lateral variations.

In this paper, we propose \core, a robust place recognition system based on a low-cost single-chip radar. As shown in Fig.~\ref{fig:place_regcognition}, \core identifies previously visited locations from a pre-built map database based on the single-chip radar. 
To begin with, \core sets out from the IF signal, which is the raw data of single-chip radar for generating the point cloud. Specifically, \core employs heatmap generation and feature extraction on the IF signal. Heatmap generation estimates range and angle to create the range azimuth heatmap, while feature extraction generates the place descriptor by applying a spatial encoder on the heatmap. 
Subsequently, \core compares the similarity between the generated place descriptor and the place descriptors in the pre-built map database to recognize the current location.
Moreover, \core proposes to employ a rotating platform and concatenates heatmaps in a rotation cycle. This method not only effectively compensates for the antenna gains but also significantly enhances the FOV of the single-chip radar place recognition system, enhancing the system's performance in scenarios with rotational and lateral variations.

Although there are some datasets available for the IF signal~\cite{rebut2022raw, gao2020ramp} of single-chip radar \textit{or} for the place recognition~\cite{barnes2020oxford, caesar2020nuscenes}, we are unable to find a dataset specifically focusing on the IF signal of single-chip radar for place recognition. As a result, we establish a dataset named milliSonic for single-chip radar place recognition, incorporating data from the USTC campus, the city roads surrounding the campus, and an underground parking garage. Subsequently, we conduct a performance evaluation of \core on this self-collected milliSonic dataset. 

\begin{figure*}[t]
    \vspace{5pt}
    \centering
    \setlength{\abovecaptionskip}{0.cm}
    \includegraphics[width=0.95\linewidth]{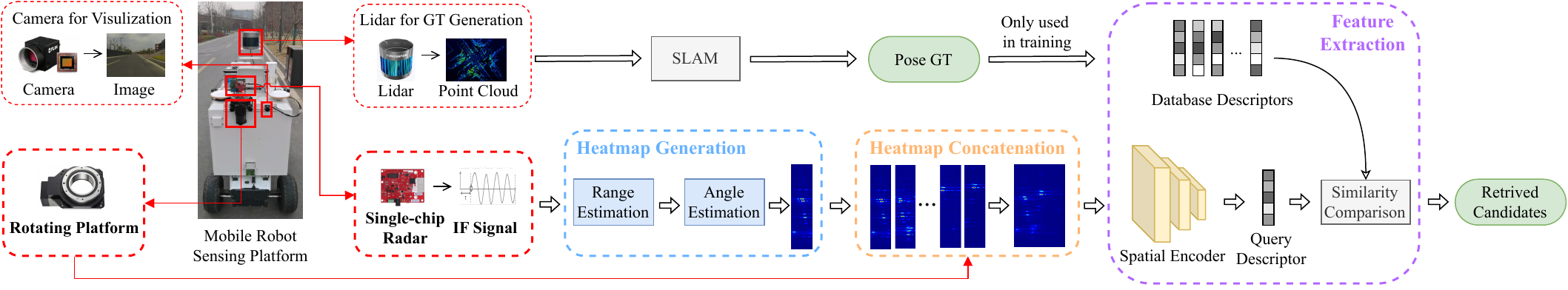}
    \caption{The \core system consists of three main components: (1) heatmap generation, (2) feature extraction, and (3) heatmap concatenation. The heatmap generation module performs range estimation and angle estimation on the IF signal to generate the range azimuth heatmap. The feature extraction module employs a spatial encoder on the heatmap to produce the place descriptor. The heatmap concatenation module employs a rotating platform and concatenates the heatmaps over a full rotation cycle.}
    \label{fig:pipeline}
    \vspace{-15pt}
\end{figure*}

In summary, our contributions are as follows:
\begin{itemize}
\item
We propose \core, a robust place recognition system based on a low-cost single-chip radar. This system transforms the  IF signal into the range azimuth heatmap and employs a spatial encoder to extract heatmap features for place recognition, outperforming both point cloud-based and point cloud image-based methods.
\item
Our \core proposes to employ a rotating platform and concatenate heatmaps based on the relative pose in a rotation cycle, achieving 87.37\% recall@1 in scenarios encompassing rotational variations from 0 to 40 degrees and lateral translation variations from 0 to 3 meters.
\item
We collect a dataset, called milliSonic\footnote{The dataset and its related code are released here: https://github.com/TC-MCZ/mmPlace}, for the single-chip radar place recognition on the USTC campus, the city roads surrounding the campus, and an underground parking garage.
\end{itemize} 

\section{RELATED WORK}
In this section, we review the related work on place recognition, which can be classified into three categories based on the sensors: camera, LiDAR, and radar.

\subsection{Visual Place Recognition}
Visual place recognition is the most investigated technique in the place recognition area because of the camera's ubiquity, rich information, and cost-effectiveness. Initially, Cummins et al.~\cite{cummins2008fab} propose a non-learning method called FAB-MAP that uses SIFT features to construct a Bag-of-visual-words (Bow) architecture for place recognition. However, learning-based place recognition methods have become more prominent with the advent of learning-based feature extraction~\cite{SimonyanZ14a,namatevs2017deep}. For example, Arandjelovic et al.~\cite{arandjelovic2016netvlad} introduce NetVLAD, a generalized VLAD layer that enhances the generalization ability of visual place recognition. Building upon NetVLAD, Hausler et al.~\cite{hausler2021patch} develop Patch-NetVLAD, which combines the benefits of local and global descriptor techniques by deriving patch-level features from the VLAD layer that are highly invariant to translation and rotation changes. 

\subsection{LiDAR-based Place Recognition}
Unlike cameras, LiDAR can capture the 3D geometric structure of the surrounding environment using laser beams. Kim et al.~\cite{kim2018scan} propose ScanContext, a rotation-invariant 3D place descriptor that directly records the 3D structure of visible space. As an extension of ScanContext, Kim et al.~\cite{kim2021scan} develop ScanContext++, a generic descriptor that is robust to both rotation and translation. With the emergence of learning-based 3D feature extraction, Vidanapathirana et al.~\cite{vidanapathirana2022logg3d} introduce LoGG3D-Net, which employs a local consistency loss to guide the network to learn the consistent local features.

\subsection{Radar-based Place Recognition}
\label{sec:radar_place_recognition}
Unlike cameras and LiDAR, radar operates at much lower frequencies (GHz), making them insensitive to degraded visual environments such as rain, dust, fog, and direct sunlight. Hong et al.~\cite{hong2020radarslam} develop RadarSLAM by directly employing the LiDAR place recognition method M2DP~\cite{he2016m2dp}. Barnes et al.~\cite{barnes2020under} introduce a self-supervised framework for odometry estimation and use an intermediate feature as a global descriptor for place recognition. Cautoplai et al.~\cite{cai2021autoplace} propose AutoPlace, which extracts spatial-temporal features from the radar point cloud for place recognition, utilizing five single-chip radars. These methods, which utilize the point cloud, involve two types of input: one directly using the point cloud and the other projecting it onto a fixed-size bird's-eye-view image (called point cloud image). These methods are effective with mechanical radar such as CTS350-X, or when using five single-chip radars like ARS408. However, their performance significantly degrades when applied to a low-cost single-chip radar, such as AWR1642, which is the single-chip radar utilized in our \core and is generally more budget-friendly (as shown in Tab.~\ref{tab:table_radar_type}).

\section{METHOD}

The overview of \core is illustrated in Fig.~\ref{fig:pipeline}. The single-chip radar provides the raw data IF signal as input to \core. The heatmap generation module performs range estimation and angle estimation on the IF signal to generate the range azimuth heatmap, which is discussed in detail in Sec.~\ref{sec:heatmap_genertion}. Next, the feature extraction module employs a spatial encoder on the heatmap to produce the place descriptor, which is explained in Sec.~\ref{sec:feature_extraction}. Subsequently, \core compares the similarity between the place descriptor and the descriptors in the pre-built map database to recognize the current place. Additionally, \core deploys a rotating platform and concatenates the heatmaps over a full rotation cycle, as described in Sec.~\ref{sec:heatmap_concatenation}.

\begin{figure}[t]
\vspace{5pt}
	\centering
    \setlength{\abovecaptionskip}{0.cm}
	\includegraphics[width=0.9\linewidth]{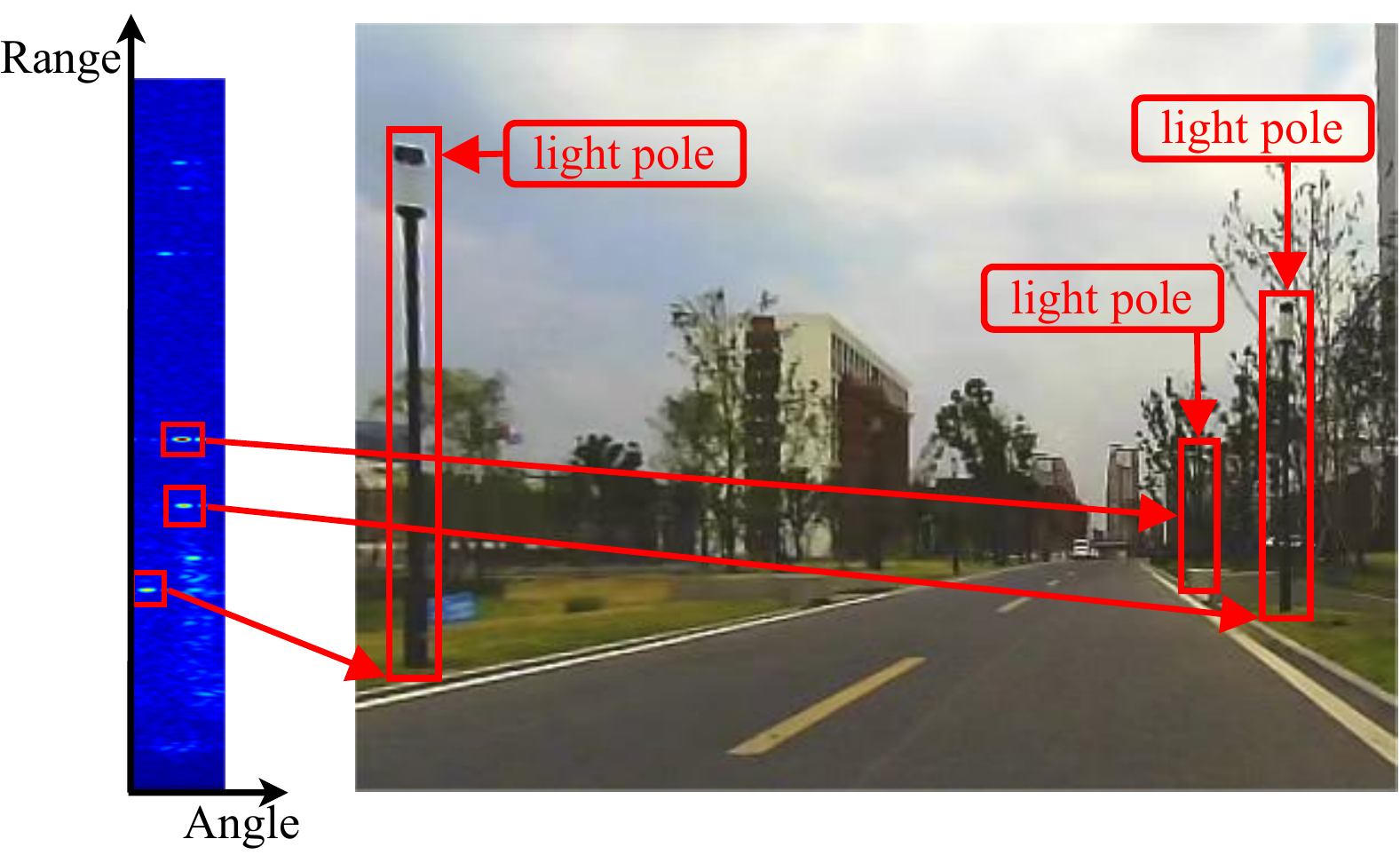}
	\caption{Visualization of the range azimuth heatmap. The heatmap shows the location of objects and indicates the power of the signal reflected back from them. As an example, the bright spots in the heatmap correspond to the light poles seen in the photo.}
	\label{fig:heatmap_fig}
\vspace{-15pt}
\end{figure}

\subsection{Range Azimuth Heatmap Generation} 
\label{sec:heatmap_genertion}

The IF signal is the raw data of the single-chip radar for generating the radar point cloud~\cite{iovescu2017fundamentals}. However, it is important to note that the IF signal lacks spatial information and is unsuitable for subsequent feature extraction. Additionally, the sparsity of the single-chip radar point cloud also hinders its utility in feature extraction. Therefore, \core performs sequential range and angle estimations on the IF signal to generate the range azimuth heatmap~\cite{kim2020low}, which contains richer information than the point cloud and encompasses spatial details not present in the IF signal.

\vspace{3pt}\noindent\textbf{Range Estimation}. The distance $d$ between the object and the single-chip radar can be obtained through the range estimation~\cite{li2021signal}. The formula for calculating the distance $d$ is given by:
\begin{equation} \footnotesize
\label{equ:range_equ}
f_{IF} = S \tau = \frac{S2d}{c} \Rightarrow d = \frac{f_{IF}c}{2S},
\end{equation}
where $f_{IF}$ is the frequency of the IF signal, $S$ is the slope of the millimeter wave chirp frequency change, $\tau$ is the round trip time, and $c$ is the speed of light.

\vspace{3pt}\noindent\textbf{Angle Estimation}. The angle of arrival (AoA) $\theta$ between the object and the single-chip radar can be determined through the angle estimation~\cite{li2021signal}. The formula for calculating the AoA $\theta$ is given by:
\begin{equation} \footnotesize
\label{equ:angle_equ}
\omega  = \frac{2 \pi \bigtriangleup d}{\lambda} = \frac{2 \pi l \sin \left ( \theta \right ) }{\lambda } \Rightarrow \theta = \sin^{-1} \left ( \frac{\lambda\omega}{2 \pi l} \right ),
\end{equation}
where $\omega$ is the phase difference, $l$ is the distance between the two receiving antennas, and $\lambda$ is the wavelength.

Our \core sequentially performs range estimation and angle estimation on the sampled IF signal ($I$). The formula for calculating the range azimuth heatmap ($H$) is as follows:
\begin{equation} \footnotesize
\label{equ:heatmap_equ}
\begin{aligned}
H=\left | \sum_{j=1}^{N_C}(\mathop{F}\limits_{k=1}^{N_R}(\mathop{F}\limits_{i=1}^{N_S}(I(i,j,k)))) \right |,\\ 
I\in C^{N_S\times N_C\times N_R}, H\in R^{N_S\times N_R},
\end{aligned}
\end{equation}
where $F(\cdot)$ represents the Fast Fourier Transform (FFT), $N_S$ is the analog-to-digital (ADC) sampling rate, $N_C$ is the number of chirps and $N_R$ is the number of antennas. Additionally, the range azimuth heatmap ($H$) can be resized by cropping the $N_S$ dimension and interpolating the zeros in the $N_R$ dimension of the IF signal ($I$).

The range azimuth heatmap encompasses important spatial data. As illustrated in Fig.~\ref{fig:heatmap_fig}, each cell within the heatmap represents the strength of the reflected signal from an object at a particular range and angle. Much like the visual image, the range azimuth heatmap possesses the capability to detect objects such as light poles.

\subsection{Spatial Encoder-Based Feature Extraction}
\label{sec:feature_extraction}

\begin{figure}[t]
\vspace{5pt}
    \centering
    \setlength{\abovecaptionskip}{0.cm}
    \includegraphics[width=0.9\linewidth]{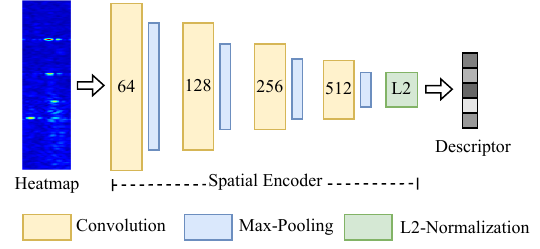}
    \caption{Overview of the spatial encoder. After obtaining the range azimuth heatmap, the spatial encoder performs feature extraction on the heatmap to obtain the place descriptor.}
    \label{fig:spatial_encoder}
\vspace{-15pt}
\end{figure}

Taking into account the similarity between the range azimuth heatmap and the Bird's Eye View image, \core employs a spatial encoder to perform feature extraction from the heatmap, resulting in the generation of the place descriptor. The spatial encoder is shown in Fig.~\ref{fig:spatial_encoder}. Initially, the heatmap consists of two dimensions: range and angle, and the convolution kernel is well-suited for extracting information from this two-dimensional space. Therefore, \core uses four layers of the convolution kernels with a ($3\times3$) kernel size and a ($1\times1$) stride to extract information from the heatmap. Subsequently, for the extraction and synthesis of relevant information, \core utilizes a max-pooling kernel after the convolution kernel layer. Given the distinct range and angle resolutions in the range azimuth heatmap, max-pooling kernels with different sizes are applied: $4$ in the range dimension and $2$ in the angle dimension.

After performing feature extraction on the range azimuth heatmap and obtaining the place descriptor, \core compares the similarity between the generated place descriptor and the descriptors stored in the pre-built map database to recognize the current place. For enhanced retrieval speed within the map database, we utilize the Faiss~\cite{johnson2019billion} library.

In addition, we perform supervised training on the spatial encoder, utilizing ground truth data from the state-of-the-art LiDAR SLAM~\cite{duan2022pfilter}. In this training process, we utilize the triplet margin loss~\cite{wang2014learning} to train the spatial encoder. This ensures that locations closer in the real world are also closer in the feature space, while locations further away in the real world are correspondingly distant in the feature space. The calculation of the loss function is as follows:
\begin{equation} \footnotesize
\label{equ:loss_equ}
L = \sum_{j}l(\min_i(d_{E}(f(q),f(p^{i})))-d_{E}(f(q),f(n^j))+\alpha),
\end{equation}
where $l(\cdot )$ is the hinge loss($l(x) = max(x, 0)$), $d_E(\cdot )$ is the Euclidean distance, $p^i$ represents the positive samples, and $n^j$ represents the negative samples.

\subsection{Heatmap Concatenation within a Rotation Cycle}
\label{sec:heatmap_concatenation}

Accurate place recognition is a challenging task for single-chip radar in scenarios with rotational and lateral variations. This challenge arises from the variability of antenna gains in different directions and the limited FOV of the single-chip radar. In particular, as the orientation of the single-chip radar changes, there are fluctuations in the signal strength reflected back from an object. In addition, there is only a limited FOV overlap of the single-chip radar between the current query data and the candidate data stored in the pre-built map database in scenarios with rotational and lateral variations. As a result, the performance of the single-chip radar place recognition experiences a noticeable degradation under such conditions, as shown in Tab.~\ref{tab:table_lat_rot}.

\begin{figure}[t]
\vspace{5pt}
	\centering
    \setlength{\abovecaptionskip}{0.cm}
	\includegraphics[width=0.95\linewidth]{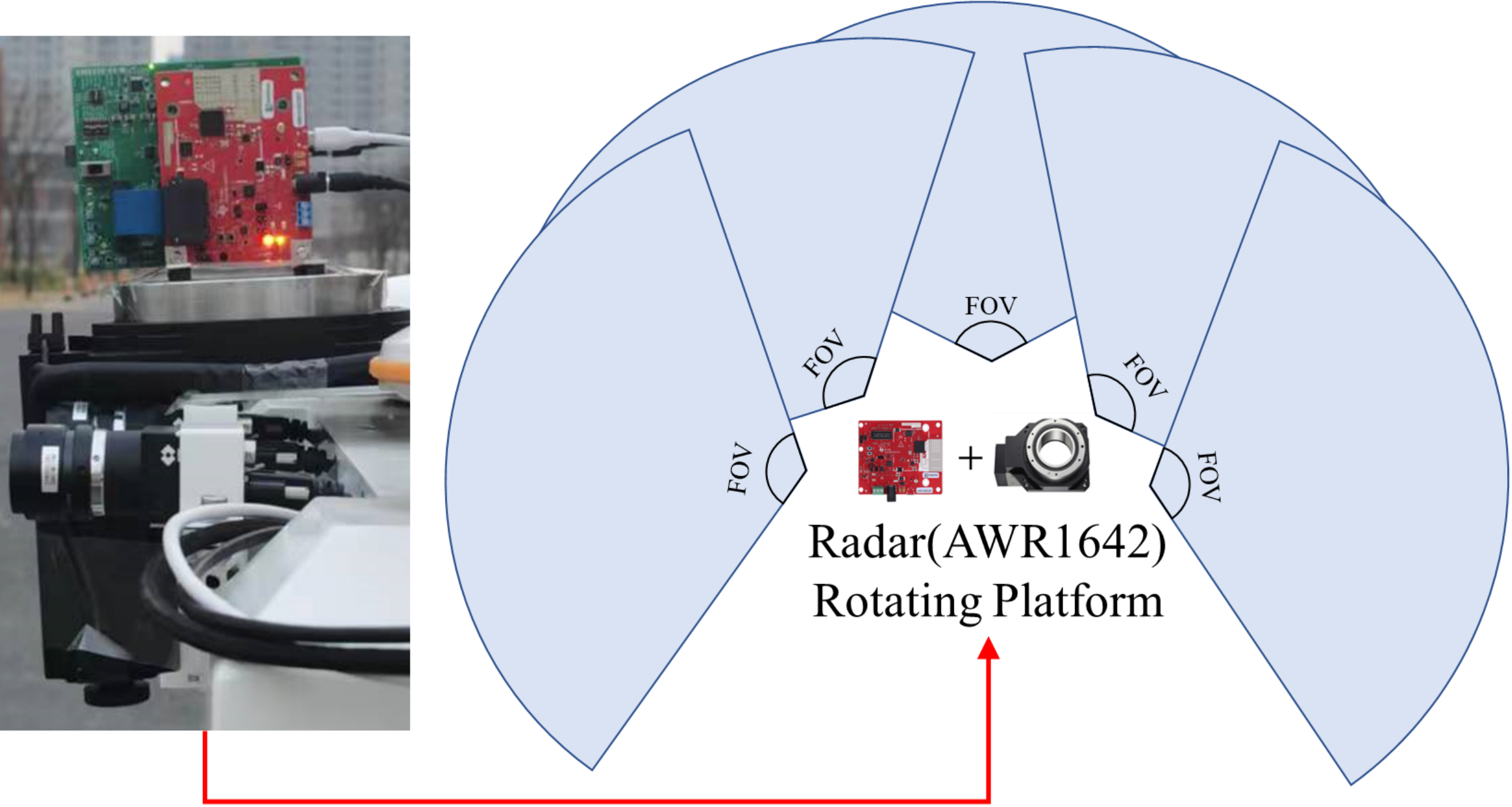}
	\caption{The radar is deployed on a rotating platform. The rotating platform rotates horizontally over 180 degrees at a speed of 150 degrees per second, which effectively captures data from multiple angles.}
	\label{fig:rotating_platform}
\vspace{-10pt}
\end{figure}

To address this challenge, as shown in Fig.~\ref{fig:rotating_platform},  we propose to employ a rotating platform to capture single-chip radar data from varying angles and concatenate the heatmaps from a complete rotation cycle. For instance, if the rotational velocity is 150 degrees per second, and the frame rate of the single-chip radar is 10 Hz, we have 12 heatmaps within a rotation cycle. Therefore, the heatmap concatenation not only can effectively compensate for antenna gains but also significantly enhances the FOV of the single-chip radar place recognition system, expanding it to 300 degrees. Thus, \core also has a larger FOV overlap in scenarios involving rotational and lateral variations.

\begin{figure}[t]
	\centering
    \setlength{\abovecaptionskip}{0.cm}
	\includegraphics[width=0.95\linewidth]{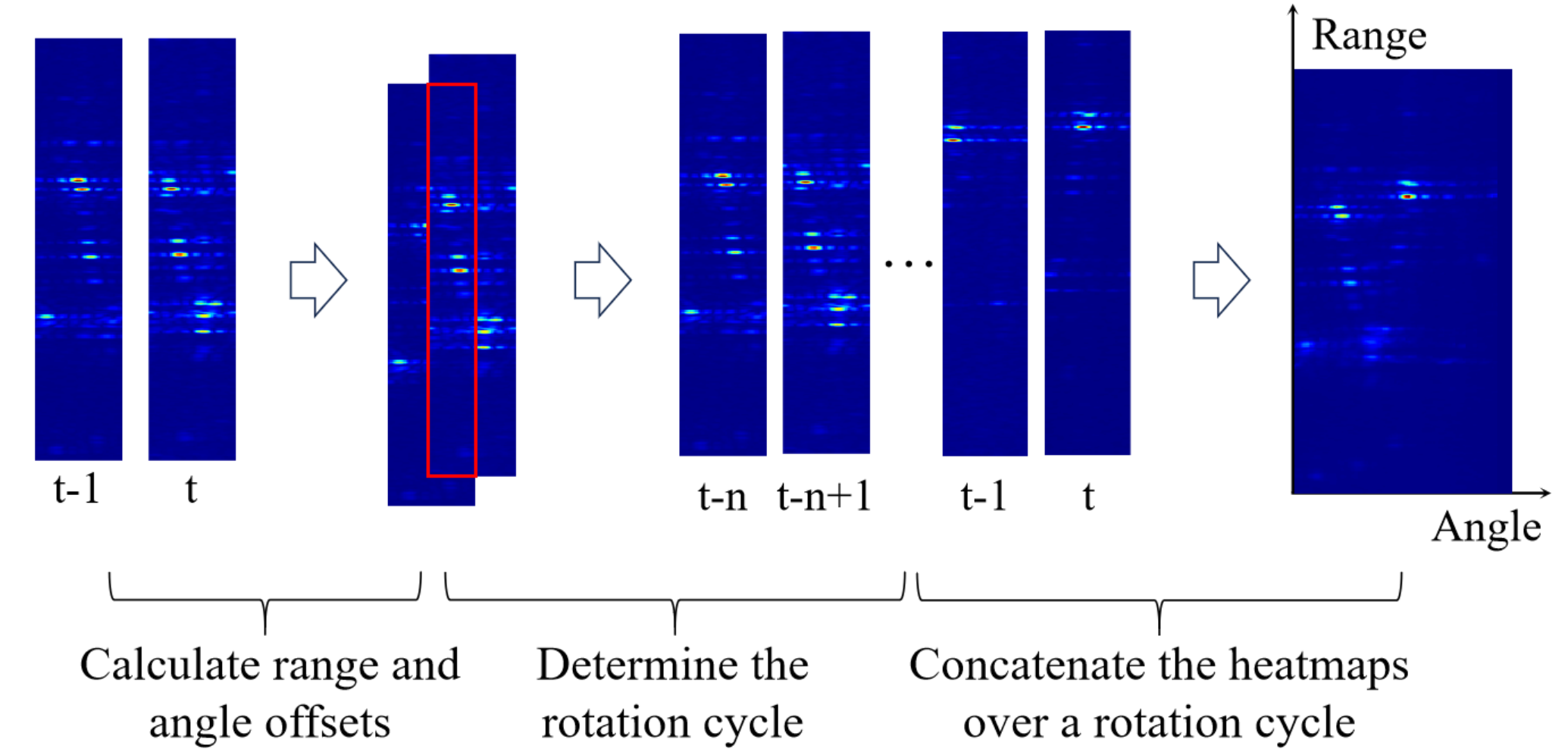}
	\caption{The process of range azimuth heatmap concatenation. Our \core concatenates heatmaps across a rotation cycle, considering both range and angle offsets. This results in a concatenated heatmap with an expanded FOV of up to 300 degrees.}
	\label{fig:heatmap_joint}
\vspace{-15pt}
\end{figure}

Next, we discuss how to concatenate the range azimuth heatmaps within a cycle, which requires precise pixel-level alignment.  The rotating platform doesn't consistently maintain a constant speed because it undergoes periodic starts and stops during the rotation. As a result, directly concatenating heatmaps at fixed step intervals can lead to a significant accumulation of concatenation errors. Therefore, \core concatenates the heatmaps based on the relative pose between neighboring frames, effectively eliminating the cumulative concatenation errors. This method takes into consideration both the range offset ($r_{offset}$) and the angle offset ($a_{offset}$) for improved accuracy. The process of heatmap concatenation is shown in Fig.~\ref{fig:heatmap_joint}. 
To begin with, we calculate the range and angle offsets between neighboring heatmap frames by minimizing the cosine similarity in their overlapping regions. The formula is expressed as follows:
\begin{equation} \small
\label{equ:cal_shifts}
r_{{offset}_{[t]}}, a_{{offset}_{[t]}} = \mathop{\arg\min}_{r, a} s(o(H_{t-1}, H_{t}(r, a))),
\end{equation}
where $H$ is the range azimuth heatmap, $H(r, a)$ means $H$ translates $r$ and $a$ pixels along the horizontal axis and the vertical axis respectively, $o(\cdot)$ represents the overlap area of $H_{t-1}$ and $H_t(r, a)$, $s(\cdot)$ computes the cosine similarity of the overlap area of $H_{t-1}$ and $H_t(r, a)$, and $r_{{offset}_{[t]}}\text{ and } a_{{offset}_{[t]}}$ are the estimated results.

Secondly, the angle offset ($a_{offset}$) possesses a positive value when the rotating platform undergoes clockwise rotation and a negative value during counterclockwise rotation. Therefore, by assessing the sign of the angular offset, we can ascertain the direction of rotation in the current frame. This enables us to identify a complete rotation cycle, where the angular offset's sign remains constant throughout that cycle. The formula is presented as follows:
\begin{equation}  \small
a_{offset_{[i-n]}} \oplus a_{offset_{[i-n+1]}}\oplus\cdots\oplus a_{offset_{[i]}} = 0,
\end{equation}
where $n$ represents the number of heatmaps for a full rotation cycle, and $\oplus$ is used to determine whether the signs of the operands on both sides of $\oplus$ are equal. If they are equal, the result is 0; if they differ, the result is 1. When conducting chained calculations, determine whether the signs of all operands involved in the computation are equal.

Finally, considering both the range and angle offsets, we concatenate the heatmap for the entire rotation cycle in a clockwise direction. The formula is presented as follows:
\begin{equation} \small
H^{'}=\bigcup_{i=1}^{n}H_i(\sum_0^i r_{{offset}_{[t]}}, \sum_0^i a_{{offset}_{[t]}}), 
\end{equation}
The resulting heatmap ($H^{'}$), derived from the concatenation of multiple angles, not only compensates for antenna gains but also extends the system's FOV, thereby significantly enhancing the system's performance in scenarios involving rotational and lateral variations. 

It is worth noting that the rotating platform is quite affordable at \$90. As shown in Tab.~\ref{tab:table_radar_type}, the rotating platform is significantly cheaper than a mechanical millimeter-wave radar and costs less than one-third of AWR1642, the single-chip radar used in \core.

\begin{table}[t]

\vspace{5pt}

\renewcommand\arraystretch{1}  
\setlength{\tabcolsep}{1.5mm}  

\centering

\setlength{\abovecaptionskip}{0cm}
\caption{Statistics of milliSonic Dataset}
\label{tab:table_milliSonic}

\begin{threeparttable}
\begin{tabular}{ccccccc}
\toprule
Seq & Frame & Len$(m)$ & Rot$(^{\circ})$ & Lat$(m)$ & Area    & Other \\ \midrule
0   & 23201 & 3200.34  & 0-10            & 0-1      & campus  & w/o RP\tnote{1}\\ 
1   & 20528 & 2706.96  & 0-10            & 0-1      & campus  & w/o RP\tnote{1}\\
2   & 31417 & 4259.45  & 0-10            & 0-1      & campus  & w/o RP\tnote{1}\\
3   & 21048 & 2756.76  & 0-10            & 0-1      & city    & w/o RP\tnote{1}\\
4   & 8253  & 809.37   & 0-10            & 0-1      & parking & w/o RP\tnote{1}\\
5   & 42351 & 3976.59  & 0-40            & 0-3      & campus  & w/o RP\tnote{1} \\ \midrule
6   & 5055  & 539.97   & 0-40            & 0-3      & campus  & w/ RP\tnote{2}  \\
7   & 12684 & 1292.33  & 0-40            & 0-3      & campus  & w/ RP\tnote{2}  \\
8   & 11911 & 764.24   & 0-40            & 0-3      & parking & w/ RP\tnote{2}  \\
9   & 12018 & 425.30   & 0-40            & 0-3      & parking & w/ RP\tnote{2}  \\
\bottomrule
\end{tabular}

\begin{tablenotes}
        \footnotesize
        \item[1] w/o RP: without the rotating platform.
        \item[2] w/ RP: with the rotating platform.
\end{tablenotes}
\end{threeparttable}

\vspace{-15pt}

\end{table}
\begin{table*}[t]
\scriptsize

\renewcommand\arraystretch{1.2} 
\setlength{\tabcolsep}{1mm}  

\centering

\setlength{\abovecaptionskip}{0cm}
\caption{Performance of \textbf{\core} and \textbf{Point Cloud-based Methods} and \textbf{Point Cloud Image-based Methods}. Sequence 1 to 4 encompass rotational variations from 0 to 10 degrees and lateral translation variations from 0 to 1 meter.}
\label{tab:table_compare_if}

\begin{tabular}{ccccccccccc}
\toprule
\multirow{2}*{Input Data} & \multirow{2}*{Method} & \multicolumn{2}{c}{Sequence 1} & \multicolumn{2}{c}{Sequence 2} & \multicolumn{2}{c}{Sequence 3} & \multicolumn{2}{c}{Sequence 4} & \multirow{1}*{Mean} \\ \cmidrule(lr){3-4} \cmidrule(lr){5-6} \cmidrule(lr){7-8} \cmidrule(lr){9-10} \cmidrule(lr){11-11}
~ & & Recall@1/5/10 & max$F_1$ & Recall@1/5/10 & max$F_1$ & Recall@1/5/10 & max$F_1$ & Recall@1/5/10 & max$F_1$ &Recall@1 \\ \midrule
\multirow{3}*{Pointcloud} & M2DP~\cite{he2016m2dp}                  & 52.60/62.77/67.14 & 0.714 & 40.16/49.03/53.59 & 0.596 & 30.39/41.78/47.38 & 0.471 & 45.40/57.68/63.74 & 0.677 & 42.13 \\
~                         & LoGG3D~\cite{vidanapathirana2022logg3d} & 56.59/70.14/74.80 & 0.772 & 53.36/69.55/75.38 & 0.723 & 38.71/54.89/62.29 & 0.568 & 46.49/67.16/71.85 & 0.681 & 48.78 \\
~                         & ScanContext~\cite{kim2018scan}          & 51.46/59.80/64.12 & 0.679 & 47.98/56.85/61.07 & 0.653 & 43.61/54.00/59.37 & 0.609 & 53.13/57.91/59.86 & 0.737 & 49.04 \\ \midrule
\multirow{2}*{Pointcloud Image} & Kidnapped~\cite{suaftescu2020kidnapped} & 43.78/51.37/55.35 & 0.696 & 39.98/49.44/53.75 & 0.614 & 35.24/46.74/52.30 & 0.552 & 50.03/62.09/66.51 & 0.720 & 42.25 \\
~                               & AutoPlace~\cite{cai2021autoplace}       & 58.40/63.59/65.72 & 0.781 & 59.01/65.47/68.65 & 0.753 & 53.69/62.71/67.47 & 0.705 & 58.74/64.63/67.51 & 0.764 & 57.46 \\ \midrule
\textbf{IF signal} & \textbf{Ours} & \textbf{88.47/89.15/89.27} & \textbf{0.941} & \textbf{88.60/90.40/90.98} & \textbf{0.943} & \textbf{88.16/89.42/89.90} & \textbf{0.940} & \textbf{88.34/89.38/89.48} & \textbf{0.941} & \textbf{88.39} \\ 
\bottomrule
\end{tabular}


\end{table*}
\begin{table*}[t]
\scriptsize

\renewcommand\arraystretch{1.2} 
\setlength{\tabcolsep}{1.25mm}  

\centering

\setlength{\abovecaptionskip}{0cm}
\caption{Comparison of \textbf{Different Range Azimuth Heatmap Sizes}. Sequence 1 to 4 encompass rotational variations from 0 to 10 degrees and lateral translation variations from 0 to 1 meter.}
\label{tab:table_heatmap_shape}

\begin{threeparttable}
\begin{tabular}{cccccccccc}
\toprule

\multirow{2}*{Heatmap Size} & \multicolumn{2}{c}{Sequence 1} & \multicolumn{2}{c}{Sequence 2} & \multicolumn{2}{c}{Sequence 3} & \multicolumn{2}{c}{Sequence 4} & \multirow{2}*{Latency(ms)} \\ \cmidrule(lr){2-3} \cmidrule(lr){4-5} \cmidrule(lr){6-7} \cmidrule(lr){8-9}
~ & Recall@1/5/10 & max$F_1$ & Recall@1/5/10 & max$F_1$ & Recall@1/5/10 & max$F_1$ & Recall@1/5/10 & max$F_1$ ~ \\ \midrule
$32\times512$  & 86.63/88.40/88.88 & 0.932 & 85.84/88.95/89.85 & 0.924 & 85.19/88.15/89.00 & 0.922 & 86.56/87.26/87.44 & 0.932 & \textbf{17.56} \\
$32\times768$  & 88.01/88.79/89.17 & 0.939 & 87.00/89.92/90.61 & 0.939  & 87.23/89.03/89.67 & 0.939  & 86.73/88.25/89.15 & 0.934 & 19.12 \\
$32\times1024$ & 88.16/88.94/89.02 & 0.941 & 87.67/90.01/90.73 & 0.938 & 88.01/89.83/90.30 & 0.939 & 87.81/88.31/89.51 & 0.938 & 20.29 \\ \midrule
$64\times512$  & 86.84/88.31/88.74 & 0.935 & 86.33/88.95/89.80 & 0.930 & 87.49/87.63/87.72 & 0.938 & 86.57/88.63/89.34 & 0.932 & 19.27 \\
$64\times768$\tnote{*} & 88.47/\textbf{89.15}/89.27 & 0.941 & \textbf{88.60}/90.40/90.98 & \textbf{0.943} & 88.16/89.42/89.90 & 0.940 & \textbf{88.34}/\textbf{89.38}/89.48 & \textbf{0.941} & 20.14 \\
$64\times1024$ & \textbf{88.49}/89.04/\textbf{89.30} & \textbf{0.943} & 88.33/\textbf{90.64}/\textbf{91.51} & 0.942 & \textbf{88.18}/\textbf{89.62}/\textbf{90.03} & \textbf{0.941} & 88.07/89.06/\textbf{89.51} & 0.940 & 24.39 \\
\bottomrule
\end{tabular}

\begin{tablenotes}
        \scriptsize
        \item[*] denotes the range azimuth heatmap size used in \core.
\end{tablenotes}
\end{threeparttable}

\vspace{-10pt}

\end{table*}
\begin{table}[t]

\renewcommand\arraystretch{1.3}  
\setlength{\tabcolsep}{3.5mm}    

\centering

\setlength{\abovecaptionskip}{0cm}
\caption{Performance of \core \textbf{without Heatmap Concatenation} under different rotational and lateral variations.}
\label{tab:table_lat_rot}

\begin{tabu}{|c|c|c|c|c|}
\hline
\diagbox[width=8.5em]{$Lat(m)$}{$Rot(^{\circ})$}&0-5&5-10&10-20&20-40\\ \hline
0.0-0.5&93.77&86.60&65.56&39.23\\ \hline
0.5-1.0&84.09&84.05&63.27&36.40\\ \hline
1.0-2.0&73.06&70.56&60.95&35.38\\ \hline
2.0-3.0&60.83&51.08&44.67&27.78\\ \hline
\end{tabu}

\vspace{-15pt}

\end{table}

\section{EXPERIMENTS}
In this section, we present the performance evaluation results of \core.
Firstly, due to the absence of publicly available place recognition datasets that encompass the single-chip radar IF signal, we collect a dataset on the USTC campus, the city roads surrounding the campus, and an underground parking garage, called milliSonic, as presented Sec.~\ref{sec:milliSonic_dataset}.
Next, we provide the implementation details in Sec.~\ref{sec:implementation_details}.
Following that, we compare \core's performance with the point cloud-based and point cloud image-based methods in Sec.~\ref{sec:revisit_with_small_variance}.
Then we present the performance of \core utilizing heatmap concatenation in rotational and lateral variation scenarios in Sec.~\ref{sec:revisit_with_rotational_and_lateral_variance}.

\begin{figure}
\centering
    \subfloat[][Sequence0 (campus)]{\label{fig:seq0}\includegraphics[width=0.45\linewidth]{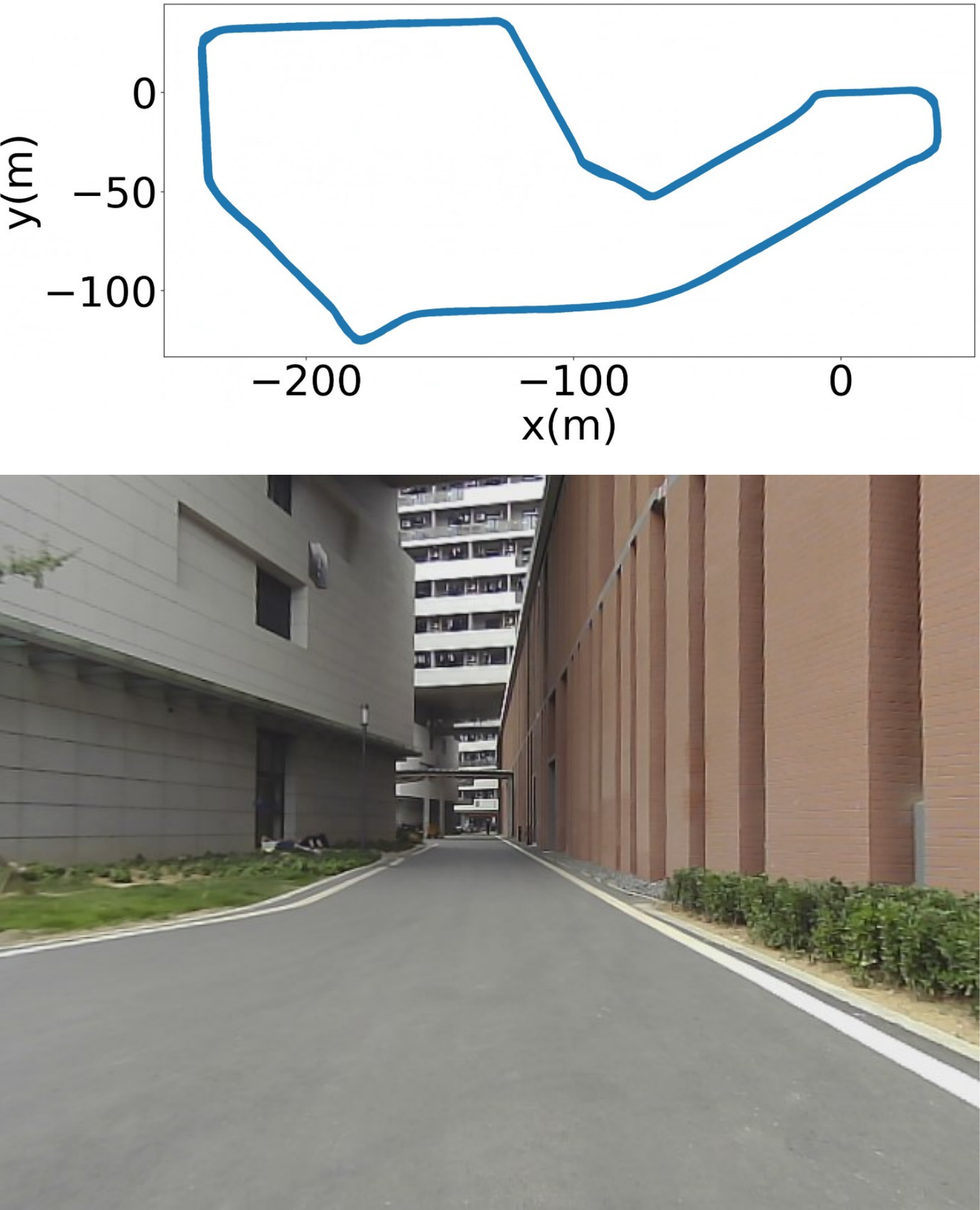}}
    \hspace{2pt}
    \subfloat[][Sequence2 (campus)]{\label{fig:seq2}\includegraphics[width=0.45\linewidth]{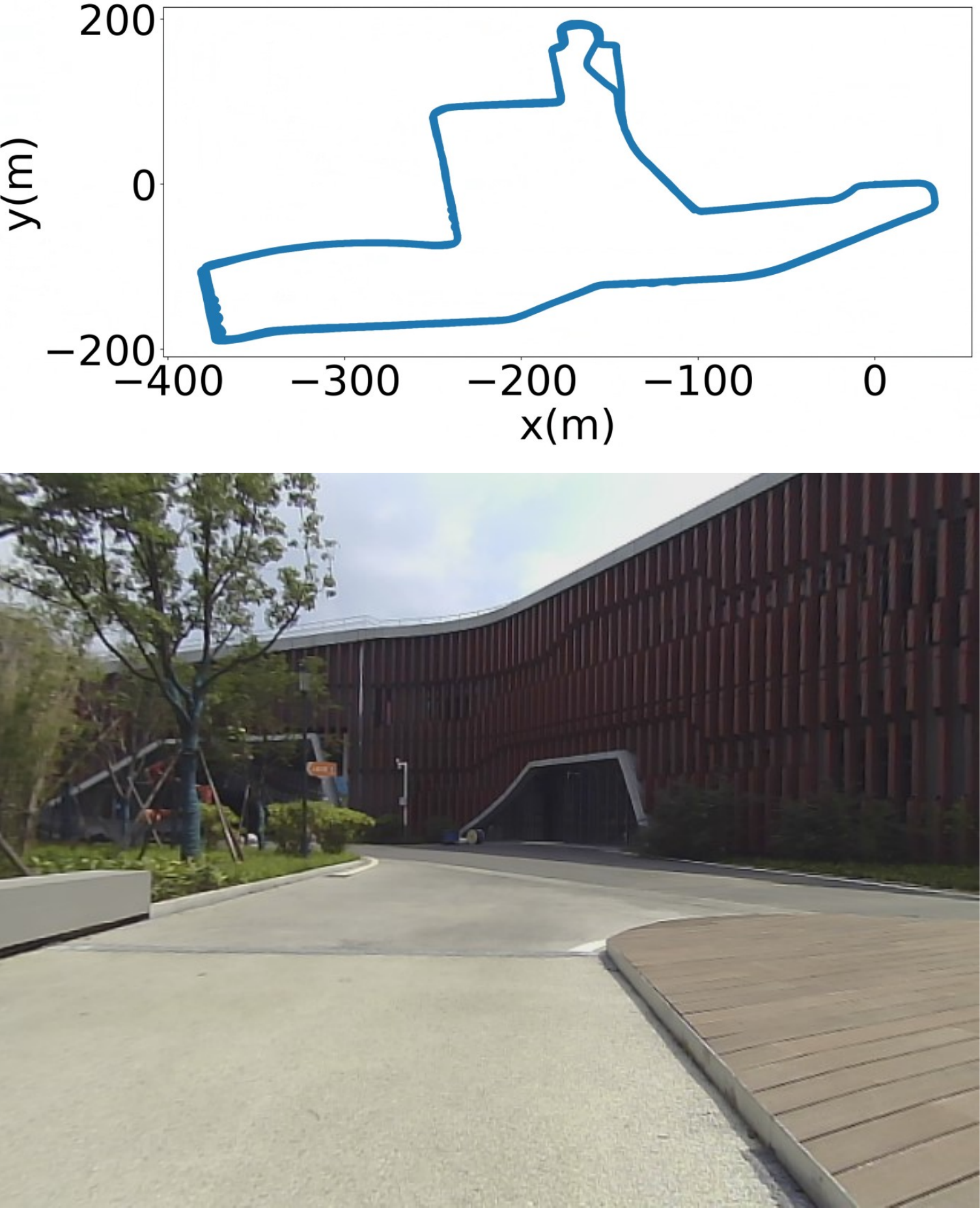}}
    \vspace{-1pt}

    \subfloat[][Sequence3 (city)]{\label{fig:seq3}\includegraphics[width=0.45\linewidth]{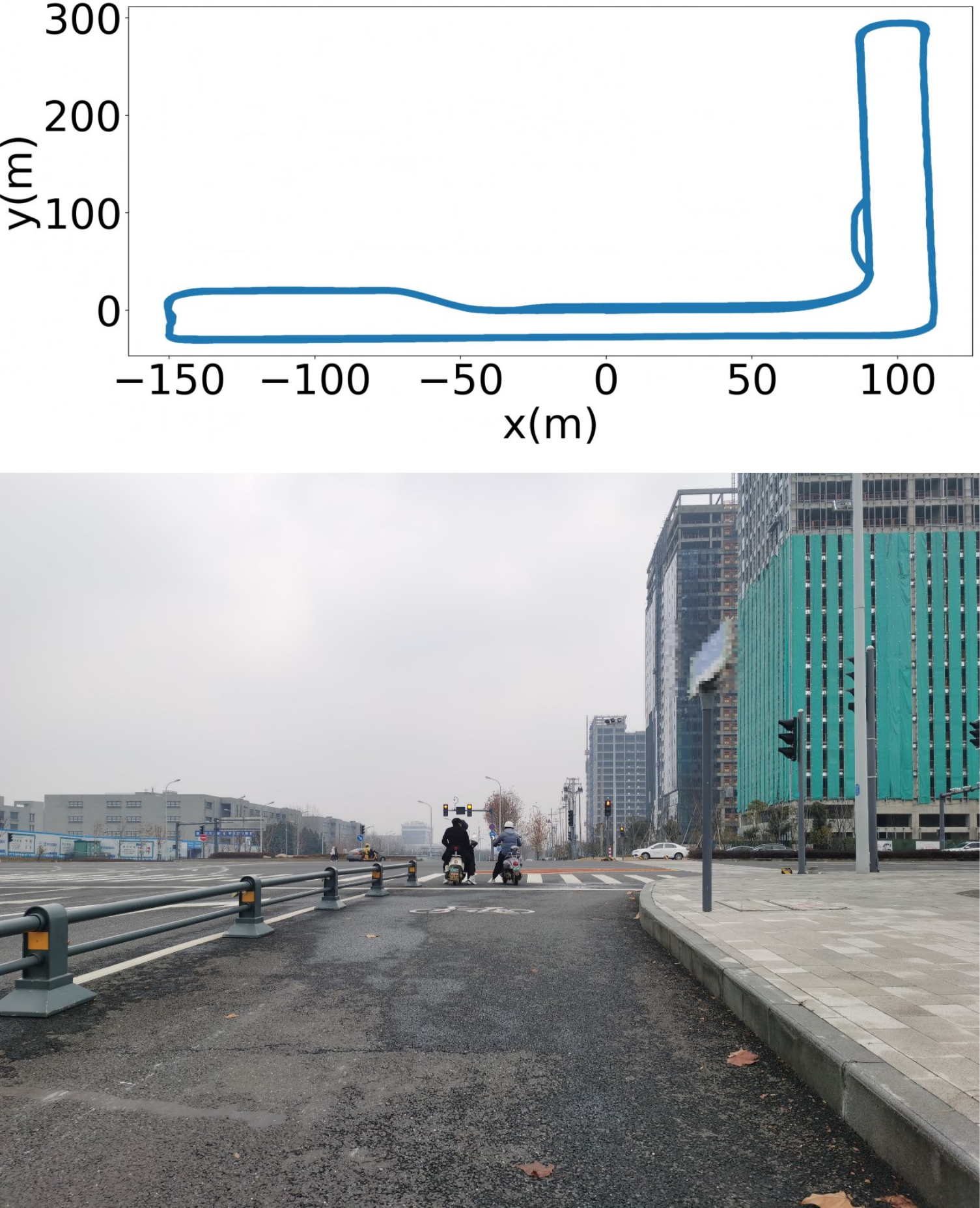}}
    \hspace{2pt}
    \subfloat[][Sequence4 (parking)]{\label{fig:seq4}\includegraphics[width=0.45\linewidth]{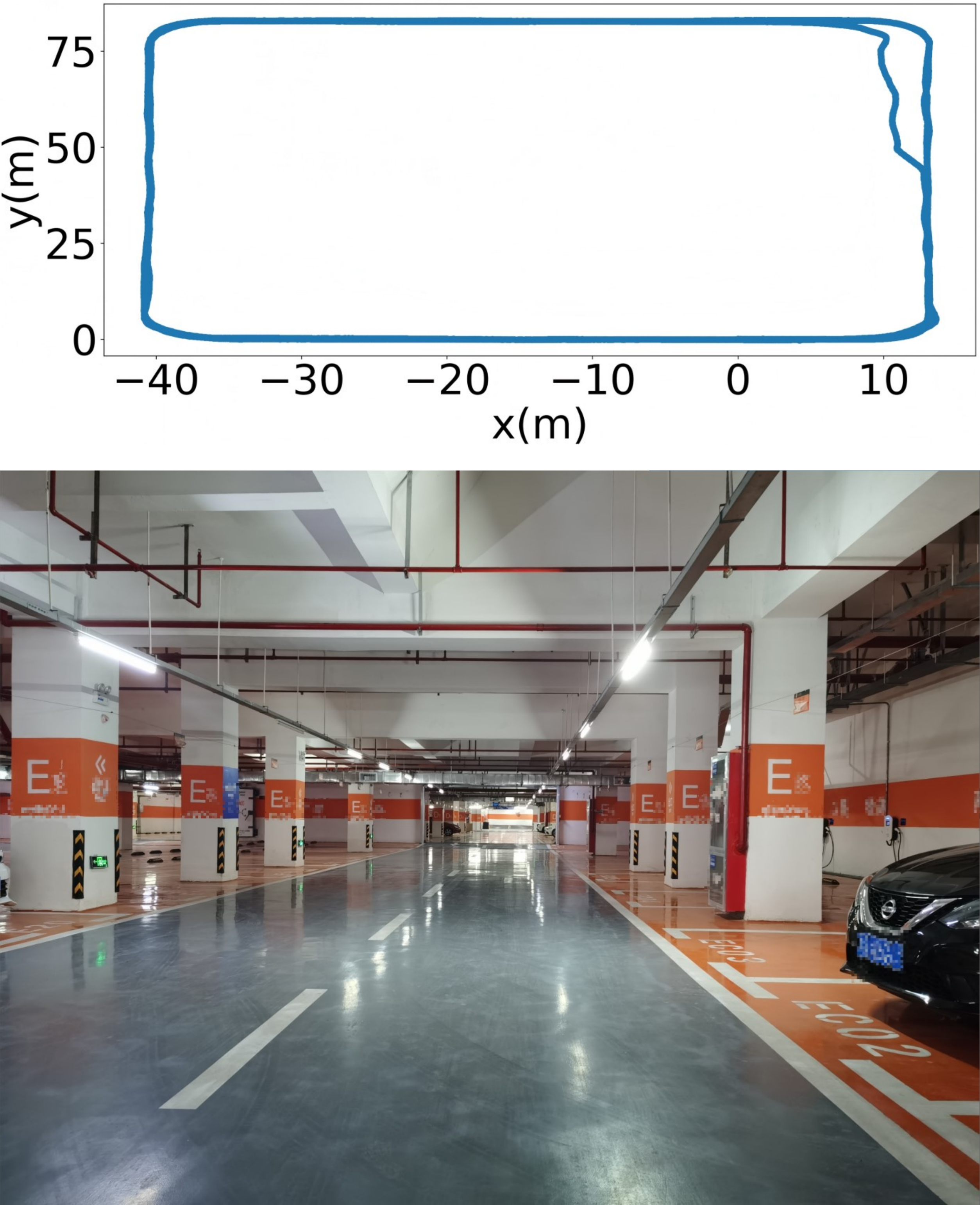}}
\caption{Data collection routes and scenarios for the four sequences in the dataset. We collect the dataset on the USTC campus, the city roads surrounding the campus, and an underground parking garage.}
\label{fig:sequences_odom}
\vspace{-15pt}
\end{figure}

\subsection{Our milliSonic Dataset}
\label{sec:milliSonic_dataset}
The existing publicly available radar datasets are not suitable for our \core. Indeed, Oxford~\cite{barnes2020oxford} and nuScenes~\cite{caesar2020nuscenes} don't include the IF signal, and RADIal~\cite{rebut2022raw} and RAMP-CNN~\cite{gao2020ramp} are designed specifically for object detection. 
Consequently, we collect the milliSonic dataset on the USTC campus, the city roads surrounding the campus, and an underground parking garage for the evaluation of \core.

For data collection, we employ a mobile robot sensing platform. As shown in Fig.~\ref{fig:pipeline}, this platform is equipped with a rotating platform, a TI AWR1642 single-chip radar sensor\footnote{https://www.ti.com/tool/AWR1642BOOST}, a LUSTER FLRA camera sensor\footnote{https://lusterinc.com/product/series1552.html}, and a RoboSense RS-32 LiDAR sensor\footnote{https://www.robosense.cn/rslidar/RS-LiDAR-32}. Thus, our milliSonic dataset includes the single-chip radar IF signal, the LiDAR point cloud, and the camera image. The LiDAR data offers precise ground truth position information via LiDAR SLAM~\cite{duan2022pfilter}, while the camera data enables observation of the robot's surroundings. Furthermore, this dataset has the potential to facilitate future research endeavors, such as multi-sensor place recognition and single-chip radar SLAM, among others.

As shown in Tab.~\ref{tab:table_milliSonic}, the milliSonic dataset comprises ten sequences, encompassing a total travel distance of 20,731 meters and comprising 188,466 frames of data. The dataset is gathered across the USTC campus, the city roads surrounding the campus, and an underground parking garage. Furthermore, as demonstrated in Fig.~\ref{fig:sequences_odom}, we present data collection routes and scenarios for the four sequences in the dataset. In the subsequent experiments, we employ sequence 0 for training, while sequence 1 to 9 are utilized for testing. Notably, data collection routes for sequence 1 and 2 exhibit a 30\% overlap with sequence 0, while routes for sequence 3 to 9 have no overlap with sequence 0. Additionally, sequence 0 to 4 encompass rotational variations from 0 to 10 degrees and lateral translation variations from 0 to 1 meter, while sequence 5 to 9 involve rotational variations from 0 to 40 degrees and lateral translation variations from 0 to 3 meters. Sequence 0 to 5 collect without the rotating platform, whereas sequence 6 to 9 utilize the rotating platform. It's worth noting that sequence 7 includes data collected both with the rotating platform (the first 6700 frames) and without the rotating platform (the last 5984 frames). 

\subsection{Implementation Details}
\label{sec:implementation_details}
The spatial encoder of \core is implemented using the PyTorch framework and trained on an Nvidia RTX 3060 GPU. For the network training, we use a batch size of 16 and employ SGD with an initial learning rate of 0.01, momentum of 0.9, and weight decay of 0.001. The learning rate is decayed by 0.5 every 5 epochs, and training continues until convergence, up to a maximum of 50 epochs. Furthermore, following the scale of AutoPlace~\cite{cai2021autoplace} and KidnappedRadar~\cite{suaftescu2020kidnapped}, we consider places in the database within a radius of 3 meters from the query as true positives, while those outside a radius of 18 meters are considered true negatives. Additionally, aligning with the AutoPlace~\cite{cai2021autoplace}, we utilize 1 positive sample and 10 negative samples for Eq.~\eqref{equ:loss_equ}.

\subsection{Evaluation in Normal Scenarios with IF Signal Data}
\label{sec:revisit_with_small_variance}
In this subsection, we evaluate \core in normal scenarios (sequence 1 to 4) involving rotational variations from 0 to 10 degrees and lateral translation variations from 0 to 1 meter. We begin by comparing \core with point cloud-based and point cloud image-based methods. Subsequently, we evaluate \core's performance under different range azimuth heatmap sizes.

The existing radar place recognition methods~\cite{he2016m2dp,kim2018scan,vidanapathirana2022logg3d, cai2021autoplace, suaftescu2020kidnapped} utilizing point cloud employ two types of input, one directly using the point cloud and the other using the point cloud image, as discussed in Sec.~\ref{sec:radar_place_recognition}. To compare with these methods that rely on point cloud or point cloud image, we also process the IF signal to generate these two types of data. Firstly, we convert the IF signal collected from the single-chip radar into 2D point clouds with the method proposed in \cite{openradar2019}. Next, we generate the pseudo-3D point clouds by adding a pseudo-axis $z=0$. Finally, we obtain the point cloud images by projecting the 2D points onto the image panel. Occupied pixels are assigned a value of 1, while unoccupied pixels are assigned a value of 0. Moreover, considering the sparse nature of the point cloud from the single-chip radar, we follow the data processing approach in ~\cite{radar:depth:20} by concatenating 5 frames of the point cloud to generate a denser point cloud.

\vspace{4pt}\noindent\textbf{IF Signal vs Point Cloud:} We compare \core with current point cloud-based~\cite{he2016m2dp,kim2018scan,vidanapathirana2022logg3d} and point cloud image-based~\cite{cai2021autoplace,suaftescu2020kidnapped} methods in sequence 1 to 4. These sequences encompass rotational variations from 0 to 10 degrees and lateral translation variations from 0 to 1 meter. The evaluation of these methods employs the standard metrics for place recognition, including recall@N~\cite{arandjelovic2016netvlad}, max$F_1$~\cite{suaftescu2020kidnapped}.

As shown in Tab.~\ref{tab:table_compare_if}, our IF signal-based \core outperforms both point cloud-based and point cloud image-based methods significantly. Specifically, \core achieves a 39.35\% and 30.93\% higher recall@1 than ScanContext (the best point cloud-based method) and AutoPlace (the best point cloud image-based method), respectively. Our \core achieves up to 88.39\% recall@1, while ScanContext and AutoPlace only achieve 49.04\% and 57.46\% recall@1, respectively. While point cloud-based and point cloud image-based methods prove effective with mechanical radars like CTS350-X or when employing five single-chip radars such as ARS408, their performance diminishes when applied to a low-cost single-chip radar. The poor performance of these methods is attributed to the sparse point cloud data of a low-cost single-chip radar, which makes it difficult to extract valid scenario information for place recognition. Whereas, \core makes full use of the information-rich IF signal data for place recognition, thus outperforming other point could-based and point cloud image-based methods significantly, as discussed in Sec.~\ref{sec:heatmap_genertion} and Sec.~\ref{sec:feature_extraction}. This also indicates that point cloud-based and point cloud image-based methods are not suitable for low-cost single-chip radar, such as AWR1642, which is the radar utilized in our \core and is generally more budget-friendly (as shown in Tab.~\ref{tab:table_radar_type}).

\begin{table*}[t]

\vspace{5pt}

\renewcommand\arraystretch{1.2} 
\setlength{\tabcolsep}{3.5mm}   

\centering

\setlength{\abovecaptionskip}{0cm}
\caption{Comparison of \textbf{Different Heatmap Concatenation Methods} in scenarios (sequence 6 to 9) involving rotational variations from 0 to 40 degrees and lateral translation variations from 0 to 3 meters.}
\label{tab:table_compare_hc}

\begin{threeparttable}
\begin{tabular}{cccccccccc}
\toprule
\multirow{2}*{Method} &\multicolumn{2}{c}{Data with 10-40 Rot$(^{\circ})$} &\multicolumn{2}{c}{Data with 1-3 Lat$(m)$} &\multicolumn{2}{c}{All Data} \\ \cmidrule(lr){2-3} \cmidrule(lr){4-5} \cmidrule(lr){6-7}
                        & Recall@1/5/10   & max$F_1$  & Recall@1/5/10   & max$F_1$  & Recall@1/5/10     & max$F_1$ \\ \midrule 
Ours  w/o HC\tnote{1}   & 65.13/68.31/69.05 & 0.823   & 68.46/74.51/75.61 & 0.839   & 69.18/76.27/80.81 & 0.840    \\ 
Ours  w/ FS-HC\tnote{2} & 79.23/80.90/82.05 & 0.906   & 77.69/82.55/83.73 & 0.881   & 82.43/84.61/85.73 & 0.915    \\ 
\textbf{Ours  w/ RP-HC\tnote{3}} & \textbf{83.48/86.42/89.88} & \textbf{0.915} & \textbf{84.47/86.84/88.84} & \textbf{0.926} & \textbf{87.37/89.07/90.92} & \textbf{0.939}   \\ \bottomrule
\end{tabular}

\begin{tablenotes}
        \footnotesize
        \item[1] w/o HC: without the heatmap concatenation. 
        \item[2] w/ FS-HC: with the fixed step-based heatmap concatenation.
        \item[3] w/ RP-HC: with the relative pose-based heatmap concatenation.
\end{tablenotes}
\end{threeparttable}

\vspace{-15pt}

\end{table*}

\begin{figure}
\centering
    \subfloat[][No Concatenation]
    {\label{fig:no_cat}\includegraphics[width=0.8\linewidth]{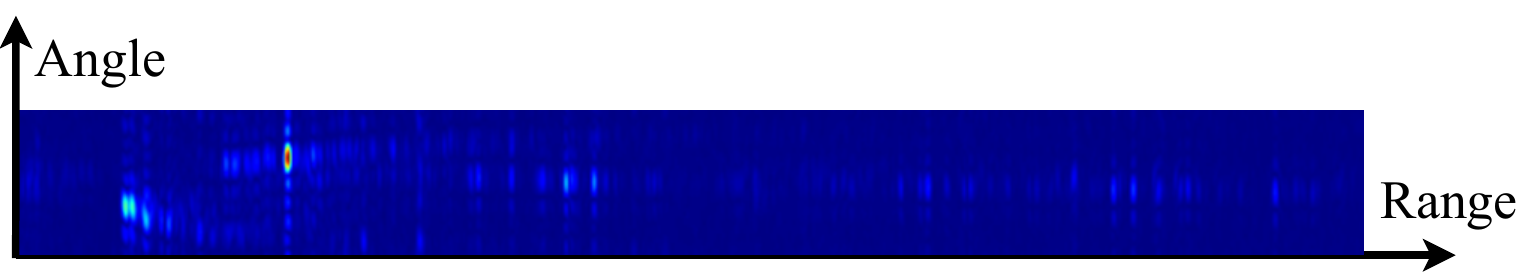}}
    \vspace{-10pt}

    \subfloat[][Fixed Step-Based Concatenation]{\label{fig:fs_cat}\includegraphics[width=0.8\linewidth]{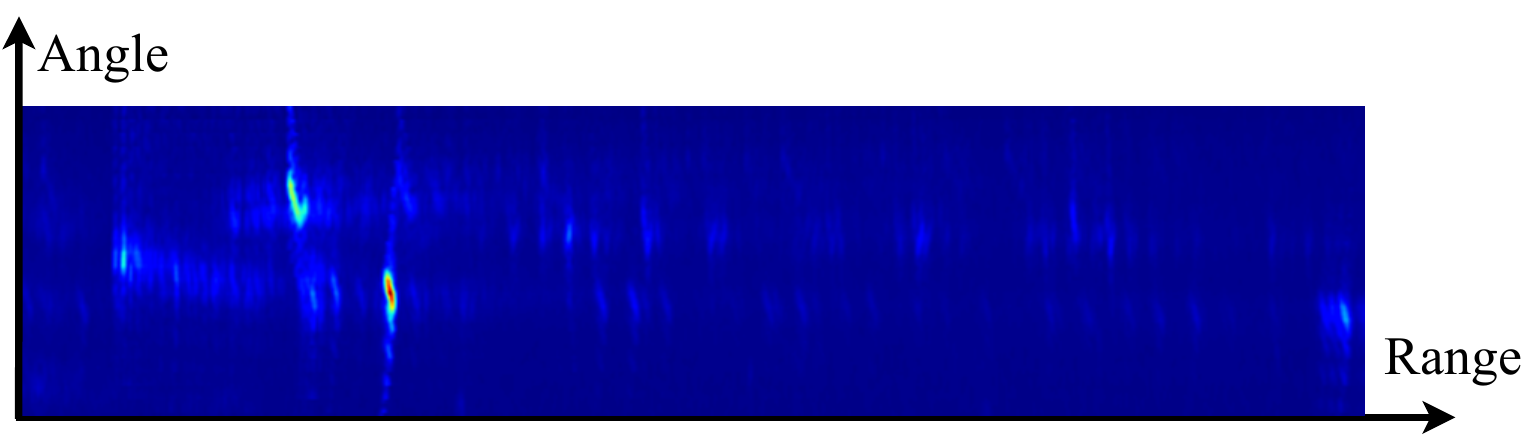}}
    \vspace{-10pt}

    \subfloat[][Relative Pose-Based Concatenation]{\label{fig:rp_cat}\includegraphics[width=0.8\linewidth]{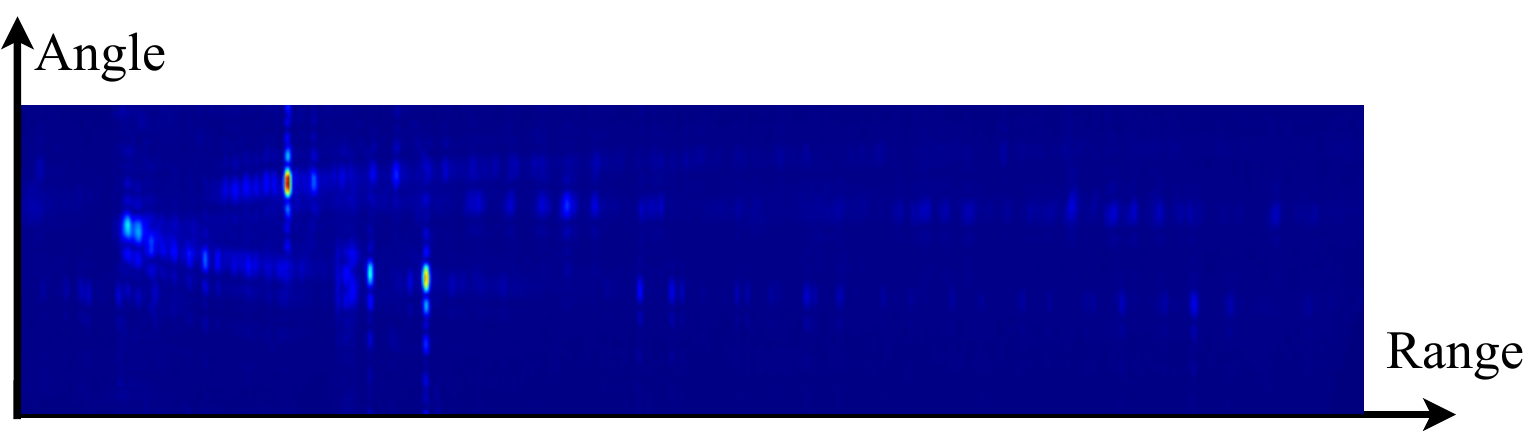}}
\caption{Visualisation of different heatmap concatenation methods. Figure (a) illustrates the heatmap without concatenation, while (b) and (c) illustrate the heatmap concatenated with fixed step size and the heatmap concatenated based on relative pose, respectively.}
\label{fig:concat_compare}
\vspace{-15pt}
\end{figure}

\vspace{4pt}\noindent\textbf{Different Heatmap Sizes:} We evaluate the effects of different range azimuth heatmap sizes in sequence 1 to 4. We modify the generated heatmap size by cropping and zero-padding on the IF signal data, as discussed in Sec.~\ref{sec:heatmap_genertion}. As shown in Tab.~\ref{tab:table_heatmap_shape}, we observe that the larger heatmap sizes result in improved performance but at the expense of increased system latency. This trade-off between performance and efficiency arises because the larger heatmap contains more information, but it also demands a higher time overhead. Notably, when enlarging the heatmap size to $64\times1024$, the improvements in performance become marginal compared to the $64\times768$ configuration, while incurring a significantly higher computational overhead. Hence, We opt for the $64\times768$ heatmap size in our \core.

\subsection{Evaluation for Heatmap Concatenation}
\label{sec:revisit_with_rotational_and_lateral_variance}
In this subsection, we evaluate \core in difficult scenarios (sequence 5 to 9) involving rotational variations from 0 to 40 degrees and lateral translation variations from 0 to 3 meters. First, we evaluate \core's performance without heatmap concatenation under different rotational and lateral variations. Afterward, we compare different heatmap concatenation methods in these difficult scenarios.

\vspace{4pt}\noindent\textbf{Different Rotational and Lateral Variations:} We evaluate the influence of different rotations and lateral translation in sequence 5, without utilizing the heatmap concatenation. This sequence encompasses rotational variations from 0 to 40 degrees and lateral translation variations from 0 to 3 meters. As shown in Tab.~\ref{tab:table_lat_rot}, the system performance declines as the rotation angle and lateral distance increase. This issue mainly arises from the limited overlap between the current query data and the candidate data stored in the pre-built map database in scenarios with rotational and lateral variations, as discussed in Sec.~\ref{sec:heatmap_concatenation}. To address this issue, we propose to employ a rotating platform and concatenate the heatmaps in a rotation cycle to enhance the system's FOV.

\vspace{4pt}\noindent\textbf{Comparison of Different Concatenation Methods:} 
We conduct a comparison of different range azimuth heatmap concatenation methods, including no concatenation, fixed step-based heatmap concatenation, and relative pose-based heatmap concatenation in sequence 6 to 9. These sequences encompass rotational variations from 0 to 40 degrees and lateral translation variations from 0 to 3 meters.

As shown in Tab.~\ref{tab:table_compare_hc}, our proposed relative pose-based heatmap concatenation achieves 87.37\% racall@1 in scenarios involving rotational and lateral variations. Specifically, it achieves 18.35\% and 16.01\% higher recall@1 than no concatenation in scenarios with 10 to 40 degrees of rotation and 1 to 3 meters of lateral translation, respectively. Also in these scenarios, our \core achieves up to 83.48\% and 84.47\% racall@1, which outperforms fixed step-based concatenation by 4.25\% and 6.78\%, respectively. This is because the relative pose-based heatmap concatenation not only enables pixel-level alignment of the heatmaps in a rotation cycle but also enlarges the FOV overlapping area in rotational and lateral variation scenarios, as discussed in Sec.~\ref{sec:heatmap_concatenation}. Again, as shown in Fig.~\ref{fig:concat_compare}, the heatmap concatenated based on relative pose has a larger FOV than the initial heatmap and is more accurate than the heatmap concatenated with fixed step size.
\section{CONCLUSION}
In this paper, we propose \core, a robust place recognition system based on a low-cost single-chip radar. Since point cloud-based and point cloud image-based methods perform poorly due to the sparse point cloud of the single-chip radar, \core starts with the IF signal. Firstly, \core generates range azimuth heatmap by performing range and angle estimation processing on the IF signal. Then, a spatial encoder is used for feature extraction on the heatmap. Additionally, \core deploys a rotating platform and concatenates heatmaps in a rotation cycle to enhance the system's performance in scenarios involving rotational and lateral variations. We collect the milliSonic dataset on the USTC campus, the city roads surrounding the campus, and an underground parking garage. Our experiments on the milliSonic dataset demonstrate that \core surpasses both point cloud-based and point cloud image-based methods. The heatmap concatenation enhances the system's performance in scenarios encompassing rotational variations from 0 to 40 degrees and lateral translation variations from 0 to 3 meters.

\addtolength{\textheight}{-12cm}  

\bibliographystyle{IEEEtran}
\bibliography{IEEEabrv, references}

\end{document}